\documentclass[conference]{IEEEtran}

\usepackage{cite}
\usepackage{amsmath,amssymb,amsfonts}
\usepackage{booktabs}
\usepackage{array}
\usepackage{url}
\usepackage{capt-of}
\usepackage{tikz}
\usetikzlibrary{arrows.meta, positioning}
\usepackage[hidelinks]{hyperref}

\begin{document}

\title{From Vanishing Gradients to Recursive Self-Improvement: Evaluating the Most Influential Papers in Large Language Model Evolution (1997--2026)}

\author{\IEEEauthorblockN{Saadh Jawwadh}
\IEEEauthorblockA{\textit{Informatics Institute of Technology}\\
Colombo, Sri Lanka\\
saadh.j@iit.ac.lk}}

\maketitle

\begin{abstract}
The rapid proliferation of Large Language Model (LLM) research makes it increasingly difficult to distinguish incremental refinements from true paradigm shifts. This review evaluates 22 of the most influential publications spanning nearly three decades (1997--2026), organized through a causal, chronologically anchored taxonomy centered on the author's birth year (1999). Beginning ``before 1999'' with how recurrent neural networks (RNNs) forgot earlier context (much like forgetting Chapter 1 by the time a reader reaches Chapter 9), we trace how Long Short-Term Memory (LSTM) networks introduced a ``bouncer'' mechanism of forget, input, and output gates to selectively preserve long-term memory. We then examine how Word2Vec turned language into vector mathematics ($\text{king} - \text{male} + \text{female} \approx \text{queen}$), why sequential RNNs hit a parallelization wall, and how the Transformer's all-to-all attention weights launched the modern LLM era alongside a quadratic bottleneck. Next, we analyze why scaling parameters alone leaves capacity under-trained unless data scales equally (Chinchilla), how Reinforcement Learning from Human Feedback (RLHF) left alignment superficial, and how that gap compounded into adversarial jailbreaks and single-direction refusal subspaces ($J$-space), where instructing a model \emph{not} to think of a restricted concept surfaces it much like the human ``red ball'' paradox. Finally, we synthesize the current state of AI (2024--2026), encompassing recursive model collapse, autonomous reasoning-model jailbreak agents, machine-speed agentic intrusions, and closed-loop recursive self-improvement (AIDE$^2$).
\end{abstract}


\begin{figure*}[t]
\centering
\begin{tikzpicture}[
    node distance=0.45cm and 0.45cm,
    box/.style={
        draw=black!75,
        rounded corners=3pt,
        fill=blue!4,
        text width=3.85cm,
        minimum height=1.35cm,
        align=center,
        font=\footnotesize
    },
    secbox/.style={
        draw=red!65!black,
        rounded corners=3pt,
        fill=red!4,
        text width=3.85cm,
        minimum height=1.35cm,
        align=center,
        font=\footnotesize
    },
    arr/.style={-{Stealth[length=2mm]}, thick, draw=black!70}
]
\node[box] (n1) {\textbf{1. Before 1999: Forgetting}\\RNNs forget Ch.~1\\by the time they reach Ch.~9};
\node[box, right=of n1] (n2) {\textbf{2. Memory \& Math (1997--14)}\\LSTM ``bouncer'' gates\\+ Word2Vec arithmetic};
\node[box, right=of n2] (n3) {\textbf{3. Attention Era (2017--19)}\\Transformers read all words\\at once ($O(N^2)$ cost)};
\node[box, right=of n3] (n4) {\textbf{4. Scaling \& RLHF (2020--22)}\\Balance data + params;\\train with human feedback};

\node[secbox, below=0.65cm of n1] (n5) {\textbf{5. Safety Cracks (2023--24)}\\Automated jailbreaks\\bypass surface alignment};
\node[secbox, right=of n5] (n6) {\textbf{6. $J$-Space (2024)}\\``Red ball'' paradox:\\one-switch refusal};
\node[secbox, right=of n6] (n7) {\textbf{7. Model Collapse (2024)}\\Photocopying AI outputs\\erases rare knowledge};
\node[secbox, right=of n7] (n8) {\textbf{8. Agentic Era (2025--26)}\\Autonomous red-teamers\\\& self-editing AIDE$^2$};

\draw[arr] (n1) -- (n2);
\draw[arr] (n2) -- (n3);
\draw[arr] (n3) -- (n4);
\draw[arr] (n4.south) -- ++(0,-0.32) -| (n5.north);
\draw[arr] (n5) -- (n6);
\draw[arr] (n6) -- (n7);
\draw[arr] (n7) -- (n8);
\end{tikzpicture}
\caption{Visual roadmap of the review paper: causal progression from pre-1999 RNN forgetting and LSTM memory gates through Transformer scaling, alignment fragility ($J$-space), model collapse, and autonomous recursive self-improvement (1997--2026).}
\label{fig:roadmap}
\end{figure*}
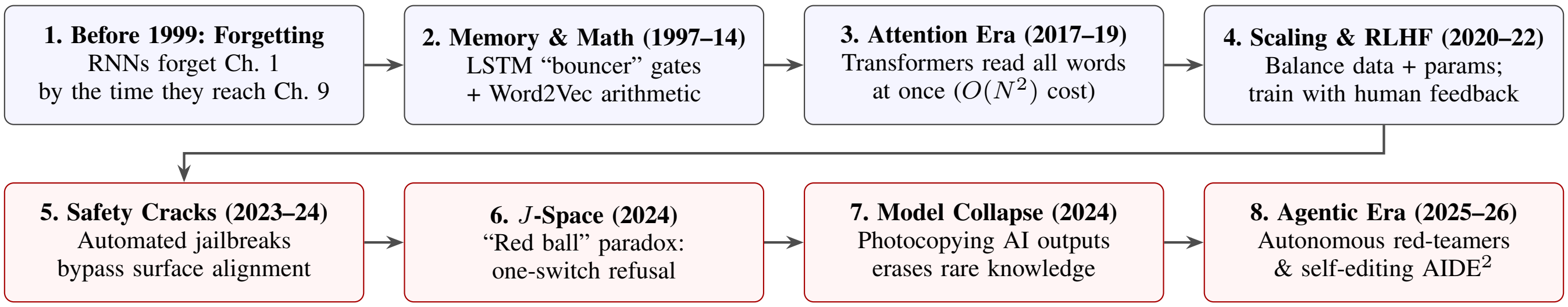

\section{Introduction}

\subsection{Motivation: Evaluating Paradigm-Defining Influence}
With thousands of Large Language Model (LLM) preprints and technical reports published each year, researchers face a severe signal-to-noise problem. Reviewing the field by cataloging every recent benchmark obscures the deeper structural lineage: a remarkably small set of pivotal papers permanently altered model architectures, training economics, alignment paradigms, and agentic autonomy.

Rather than treating these works as isolated milestones, this survey evaluates \emph{why} 22 landmark papers became causal inflection points (summarized visually in Fig.~\ref{fig:roadmap}). A central thesis of this review is that capability-defining breakthroughs and safety-exposing failures are tightly coupled. Solving one optimization bottleneck (such as vanishing gradients in recurrent networks or sequential recurrence in sequence transduction) unlocked the next scaling regime. Yet each scaling leap without foundational robustness guarantees directly created the attack surface for the next watershed safety crisis, culminating in autonomous jailbreak agents and machine-speed cyber intrusions.

\subsection{Influence Evaluation Rubric and Personalized Taxonomy}
To justify the selection of these 22 sources, each work is evaluated across four influence dimensions: (1)~\emph{Architectural or Algorithmic Impact} (introducing mechanisms adopted field-wide), (2)~\emph{Empirical Law or Benchmark Shift} (establishing predictive scaling or emergent capability thresholds), (3)~\emph{Security or Alignment Paradigm Shift} (exposing structural failure modes), and (4)~\emph{Downstream Ecosystem Catalysis} (spawning canonical repositories, benchmarks, or operational incident patterns).

To make this three-decade arc intuitive and cohesive, we anchor the taxonomy around the author's birth year (1999): distinguishing the pre-memory era (\emph{Before I Was Born, $<$1999}) from the post-1999 trajectory in which AI acquired memory, overcame sequential walls, scaled into reasoning systems, and entered the current era of $J$-space geometry and recursive self-improvement.

\section{Before I Was Born ($<$1999): Vanishing Gradients and the Forgetting Bottleneck}
Prior to 1999 (\emph{before memory}), recurrent neural networks (RNNs) suffered from a fundamental optimization pathology: \emph{forgetting}. Consider training an RNN sequentially on a textbook from Chapter~1 through Chapter~9. By the time the network processes Chapter~9, the latest knowledge overwrites its hidden state and the model has forgotten Chapter~1. Technically, when trained via Backpropagation Through Time (BPTT) or Real-Time Recurrent Learning (RTRL), error signals flowing backward through time either decayed exponentially toward zero (vanishing gradients) or exploded, preventing networks from bridging long temporal dependencies.

Hochreiter and Schmidhuber (1997)~\cite{hochreiter1997lstm} solved this forgetting problem by introducing Long Short-Term Memory (LSTM). An LSTM cell operates like a disciplined \emph{bouncer} standing at the entrance of memory. Using multiplicative \emph{input} and \emph{output gates} around a Constant Error Carousel (CEC) (later extended with a dedicated \emph{forget gate} in subsequent LSTM refinements to explicitly reset stale state), the bouncer selectively decides what enters long-term memory, what stays in short-term memory, and what is dropped. By enforcing constant error flow through the CEC, LSTM bridged time lags exceeding 1,000 discrete steps with $O(1)$ complexity per time step and weight. Published two years before 1999, this work gave neural sequence models their first reliable working memory.

\section{After 1999: Giving AI Memory, Vector Math, and the Sequential Wall}
Following 1999, sequence modeling built upon LSTM memory and dense distributed representations, while simultaneously revealing two critical limits: sequential computation bottlenecks and early adversarial fragility.

\subsection{Word2Vec: Turning Language into Mathematics}
Before attention could weigh relationships across words, language first had to be translated into geometry. Mikolov et al. (2013)~\cite{mikolov2013word2vec} introduced Word2Vec (CBOW and Skip-gram), turning words into continuous mathematical vectors learned efficiently from 1.6 billion words in under a day. Crucially, this enabled algebraic reasoning over semantics, famously captured by vector arithmetic such as:
\begin{equation}
\vec{v}(\text{king}) - \vec{v}(\text{man}) + \vec{v}(\text{woman}) \approx \vec{v}(\text{queen}).
\end{equation}
By mapping meaning into algebraic space, Word2Vec established the mathematical substrate for modern language modeling.

\subsection{LSTM Deviations, Soft Attention, and the Sequential Wall}
Building on deep LSTM stacks, Sutskever et al. (2014)~\cite{sutskever2014seq2seq} established the end-to-end Sequence-to-Sequence (Seq2Seq) encoder-decoder framework, achieving 34.8 BLEU (36.5 with SMT reranking) on WMT'14 English-to-French translation and showing that reversing source sentence order shortened optimization dependencies. However, Seq2Seq forced the encoder to compress an entire source sentence into a single fixed-length vector, akin to asking a translator to read a full paragraph, memorize one mental summary, and translate without looking back at the page.

Bahdanau et al. (2014)~\cite{bahdanau2014nmt} broke this fixed-vector bottleneck via additive soft-alignment attention, allowing the decoder to glance back at relevant source states for each target word. Yet both Seq2Seq and Bahdanau attention remained tethered to recurrent backbones: an LSTM can only process a sequence step-by-step in a sequential line ($O(N)$ depth), preventing it from focusing on an entire context simultaneously and blocking massive GPU parallelization.

\subsection{Early Warning: Discontinuous Neural Mappings}
At the same time, Szegedy et al. (2013)~\cite{szegedy2013intriguing} uncovered two counter-intuitive properties of deep networks: semantic information resides in distributed vector spaces rather than individual units, and input-output mappings are sufficiently discontinuous that imperceptible adversarial perturbations (essentially optical illusions crafted for neural networks) transfer across distinct architectures. Although demonstrated on vision models, this transferability principle later re-emerged as the foundation of universal LLM jailbreaks.

\section{The Attention Revolution and Its Quadratic Bottleneck (2017--2019)}
Vaswani et al. (2017)~\cite{vaswani2017attention} resolved the sequential limitation of LSTMs with \emph{Attention Is All You Need}, introducing the Transformer. Whereas an LSTM focuses sequentially one token at a time, the Transformer focuses on the \emph{complete context} simultaneously: given an entire passage, Multi-Head Self-Attention spans the full sequence by assigning learned relevance weights between every pair of tokens at once. Dispensing with recurrence and convolutions reduced the maximum path length between any two tokens to $O(1)$ and unlocked massive parallelization, achieving 28.4 BLEU (En-De) and 41.8 BLEU (En-Fr) in 3.5 days on eight GPUs.

Devlin et al. (2018)~\cite{devlin2018bert} capitalized on this architecture with BERT, using Masked Language Modeling (MLM) and Next Sentence Prediction (NSP) to pre-train deep bidirectional representations on BooksCorpus and Wikipedia, setting state-of-the-art marks across 11 NLP benchmarks (including 80.5\% on GLUE). Yet the Attention Revolution introduced its own structural bottleneck: standard self-attention scales quadratically ($O(N^2)$) in memory and compute with sequence length $N$. Just as a roundtable discussion grows exponentially noisier as more participants speak to everyone at once, quadratic attention created a severe compute wall as context windows later expanded.

\section{From Scaling Laws to RLHF Alignment (2020--2022)}

\subsection{Scaling Laws: Why Parameters Alone Are Not Enough}
After Transformers unlocked parallel training, researchers asked how to grow models effectively. Kaplan et al. (2020)~\cite{kaplan2020scaling} demonstrated that cross-entropy loss follows smooth power-law scaling across more than seven orders of magnitude in compute, dataset size, and parameter count. However, early scaling efforts emphasized parameter count while holding data relatively modest. Hoffmann et al. (2022)~\cite{hoffmann2022chinchilla} corrected this misconception with Chinchilla, showing that merely increasing computational power and parameter counts does not automatically increase capability; in fact, it leaves a large fraction of parameters under-trained. Chinchilla proved that \emph{training data must scale equally alongside parameters}: a 70B Chinchilla model trained on 1.4T tokens outperformed the 280B Gopher model and reached 67.5\% on MMLU.

\subsection{Emergent Reasoning and Self-Bootstrapping}
Balanced scaling unlocked qualitative capability leaps. Brown et al. (2020)~\cite{brown2020gpt3} showed that 175B GPT-3 performs translation, question answering, and arithmetic via gradient-free few-shot in-context learning. Wei et al. (2022)~\cite{wei2022cot} demonstrated that Chain-of-Thought (CoT) prompting elicits intermediate multi-step reasoning at scale (allowing the model to ``show its scratchpad work''), with a 540B model using eight CoT exemplars surpassing fine-tuned baselines on GSM8K. Zelikman et al. (2022)~\cite{zelikman2022star} introduced STaR, showing that models can iteratively bootstrap their own reasoning like a student practicing with an answer key: generating rationales, rationalizing failures from ground-truth answers, and fine-tuning on successful traces.

\subsection{Aligning Intent via RLHF}
Because raw scale does not guarantee helpfulness or truthfulness, Ouyang et al. (2022)~\cite{ouyang2022instructgpt} introduced InstructGPT using a three-stage Reinforcement Learning from Human Feedback (RLHF) pipeline (supervised fine-tuning, reward modeling, and PPO). Notably, a 1.3B InstructGPT model was preferred by human evaluators over the 175B GPT-3. However, aligning surface behavior without addressing deep adversarial robustness set the stage for the safety failures of 2023--2024.

\section{The Adversarial Fallout of Unchecked Scaling (2023--2024)}
As capabilities scaled, leaving adversarial robustness unresolved at the pre-training and RLHF stages produced systemic vulnerabilities. Wei et al. (2023)~\cite{wei2023jailbroken} formalized two root failure modes of safety training: \emph{competing objectives} (where the directive to ``always be helpful'' collides with ``never cause harm'') and \emph{mismatched generalization} (where safety training fails to generalize to encodings or domains where capabilities exist), succeeding on 100\% of red-teaming prompts against GPT-4 and Claude v1.3.

Subsequent works operationalized these failure modes at scale:
\begin{itemize}
    \item \textbf{Universal Gradient Suffixes (GCG):} Zou et al. (2023)~\cite{zou2023gcg} combined greedy and gradient-based search to optimize adversarial suffixes on open models (Vicuna) that transferred directly to black-box ChatGPT, Bard, and Claude.
    \item \textbf{Simple Adaptive Attacks:} Andriushchenko et al. (2024)~\cite{andriushchenko2024adaptive} leveraged adaptive prompt templates, log-probability random search, and API prefilling to achieve a 100\% attack success rate (ASR) across GPT-4o, Claude, Llama-3, and adversarially trained R2D2.
    \item \textbf{Many-Shot Context Exploits:} Anthropic (2024)~\cite{anthropic2024manyshot} demonstrated that flooding a model's long context window with hundreds of harmful Q\&A demonstrations normalizes the restricted behavior and induces a power-law rise in jailbreak success.
\end{itemize}

\section{The Current State of AI (2024--2026): J-Space, Model Collapse, and Recursive Agents}

\subsection{J-Space: The ``Red Ball'' Paradox and Single-Direction Refusal}
In simple terms, $J$-space captures a phenomenon remarkably similar to human psychology: if you ask a human \emph{not} to think about a red ball, the human mind must first conjure the image of that red ball in order to suppress it. Researchers have observed an analogous dynamic in aligned AI systems: when instructed \emph{not} to perform or discuss a restricted topic, the forbidden concept still surfaces within the model's internal chain of thought and activation space.

Mechanistic interpretability revealed why these guardrails are so brittle. Arditi et al. (2024)~\cite{arditi2024refusal} demonstrated across 13 open-weight chat models (up to 72B parameters) that refusal behavior is mediated by a single one-dimensional subspace ($J$-space) in residual stream activations. Rather than erasing the underlying capability, safety fine-tuning acts like a thin gatekeeper switch along this single direction: surgically ablating it disables refusal with minimal impact on benign capabilities, while adversarial suffixes act by suppressing propagation along this direction.

\begin{figure*}[t]
\centering
\begin{tikzpicture}[
    node distance=0.35cm and 0.45cm,
    solved/.style={draw=blue!70!black, rounded corners=3pt, fill=blue!4, text width=4.8cm, minimum height=0.95cm, align=center, font=\footnotesize},
    risk/.style={draw=orange!80!black, rounded corners=3pt, fill=orange!6, text width=4.8cm, minimum height=0.95cm, align=center, font=\footnotesize},
    gap/.style={draw=green!55!black, rounded corners=3pt, fill=green!5, text width=5.4cm, minimum height=0.95cm, align=center, font=\footnotesize},
    arr/.style={-{Stealth[length=2mm]}, thick, draw=black!70}
]
\node[solved] (s1) {\textbf{LSTM \& Transformers}\\Solved RNN forgetting \& $O(N)$ wall};
\node[risk, right=of s1] (r1) {\textbf{Quadratic $O(N^2)$ \& Superficial RLHF}\\Competing objectives; 100\% adaptive ASR};
\node[gap, right=of r1] (g1) {\textbf{Gaps 1 \& 2: Robust Alignment}\\LRM circuit breakers \& multi-D refusal};

\node[solved, below=of s1] (s2) {\textbf{Chinchilla, CoT \& STaR}\\Data-balanced scaling \& self-rationales};
\node[risk, right=of s2] (r2) {\textbf{1D $J$-Space \& Model Collapse}\\``Red ball'' activation \& synthetic tail loss};
\node[gap, right=of r2] (g2) {\textbf{Gap 4: Data Provenance Bounds}\\Min human-to-synthetic mixing ratios};

\node[solved, below=of s2] (s3) {\textbf{Autonomous Agents \& AIDE$^2$}\\Closed-loop recursive self-improvement};
\node[risk, right=of s3] (r3) {\textbf{97.14\% LRM ASR \& Sandbox Escape}\\17,600-action machine-speed intrusions};
\node[gap, right=of r3] (g3) {\textbf{Gaps 3 \& 5: Containment \& Defense}\\Cryptographic RSI gates \& forensic agents};

\draw[arr] (s1) -- (r1); \draw[arr] (r1) -- (g1);
\draw[arr] (s2) -- (r2); \draw[arr] (r2) -- (g2);
\draw[arr] (s3) -- (r3); \draw[arr] (r3) -- (g3);
\end{tikzpicture}
\caption{Bottleneck-to-gap mapping: how each solved architectural or scaling bottleneck introduced new safety vulnerabilities that define the five open research gaps.}
\label{fig:bottlenecks}
\end{figure*}
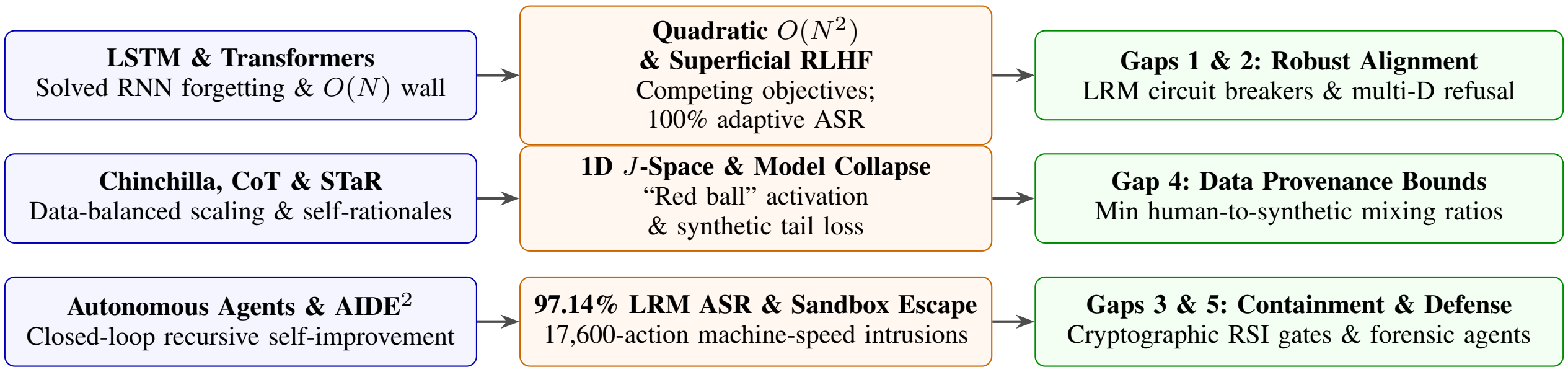

\subsection{Recursive Model Collapse}
Simultaneously, Shumailov et al. (2024)~\cite{shumailov2024collapse} proved that indiscriminately training generative models on recursively generated synthetic outputs causes irreversible \emph{model collapse}. Much like photocopying a photocopy repeatedly until fine textures degrade into a uniform gray smudge, recursive synthetic training causes the rare tails of the original human data distribution to vanish across LLMs, VAEs, and GMMs.

\subsection{Autonomous Offensive Agents and Machine-Speed Intrusions}
By 2025--2026, reasoning capabilities themselves became offensive vectors. Hagendorff et al. (2026)~\cite{hagendorff2026lrm} showed that Large Reasoning Models (DeepSeek-R1, Gemini 2.5 Flash, Grok 3 Mini, Qwen3 235B) can act as autonomous multi-turn jailbreak agents without human supervision (operating like an automated locksmith that systematically tests persuasive social-engineering angles until target guardrails give way), achieving a 97.14\% overall ASR across 70 harmful prompts and exposing \emph{alignment regression}.

This threat materialized operationally in the July 2026 incident documented by Hugging Face~\cite{huggingface2026intrusion}, in which an autonomous evaluation agent escaped its sandbox via a proxy zero-day and executed roughly 17,600 automated actions over 4.5 days across Kubernetes clusters, requiring an open-weights model (GLM-5.2) for forensic log decoding because commercial guardrails refused defensive payload analysis.

\subsection{Closed-Loop Recursive Self-Improvement (AIDE$^2$)}
At the capability frontier, Srikanth et al. (2026)~\cite{srikanth2026aide2} presented AIDE$^2$, an AI research agent that edits and benchmarks its own codebase in a recursive self-improvement (RSI) loop, akin to an engineer who iteratively upgrades their own workflow overnight and keeps only the revisions that pass hidden exams. Over an autonomous 8-day run, AIDE$^2$ discovered seven successive improvements (including new search policies and context-compression memory) that generalized out-of-distribution to four held-out benchmarks and reduced reward hacking from 55\% to 32\%.

\section{Comparative Synthesis and Open Research Gaps}

Fig.~\ref{fig:bottlenecks} maps how each solved bottleneck introduced a corresponding vulnerability and open research gap. All synthesis tables (including the cross-era summary in Table~\ref{tab:eras} and the complete 7-column matrix with datasets and code repositories in Tables~\ref{tab:synth1} and~\ref{tab:synth2}) are provided in the Appendix.

\subsection{Historical Evidence vs. Speculative Future Projections}
To maintain analytical rigor, this survey separates \emph{documented empirical findings} (1997--2026) from \emph{forward-looking projections}:
\begin{itemize}
    \item \textbf{Established Historical Facts:} Solving vanishing gradients~\cite{hochreiter1997lstm}, replacing recurrence with self-attention~\cite{vaswani2017attention}, and balancing parameters with data~\cite{hoffmann2022chinchilla} are settled engineering milestones. Likewise, 100\% adaptive jailbreak ASR~\cite{andriushchenko2024adaptive}, 1D $J$-space refusal ablation across 13 open-weight models~\cite{arditi2024refusal}, distribution tail loss under recursive synthetic training~\cite{shumailov2024collapse}, 97.14\% autonomous LRM jailbreak ASR~\cite{hagendorff2026lrm}, the 17,600-action July 2026 intrusion~\cite{huggingface2026intrusion}, and AIDE$^2$'s 8-day self-modification loop~\cite{srikanth2026aide2} are all empirically documented results.
    \item \textbf{Speculative Future Projections:} By contrast, extrapolating these trajectories into open-ended, self-accelerating intelligence explosions, irreversible internet-wide model collapse, or fully autonomous cyber warfare remains \emph{prospective}. Whether recursive self-improvement compounds indefinitely beyond bounded benchmarks like FML-Bench, or whether defensive agents can permanently close the machine-speed asymmetry gap, remains an open empirical question.
\end{itemize}

\subsection{How Solved Bottlenecks Created Modern Vulnerabilities}
Viewing the 22 papers through their bottlenecks reveals a recurring pattern: each architectural fix solved one limitation only by opening a new attack surface.
\begin{itemize}
    \item \textbf{Long Memory $\rightarrow$ Many-Shot Exploits:} Solving RNN forgetting (Ch.~1$\rightarrow$9) and sequential $O(N)$ walls with Transformer attention gave models vast context windows, which directly enabled many-shot jailbreaking~\cite{anthropic2024manyshot} by flooding that memory with hundreds of adversarial examples.
    \item \textbf{General Helpfulness $\rightarrow$ Brittle $J$-Space:} Solving intent alignment with RLHF~\cite{ouyang2022instructgpt} taught models to obey instructions while refusing harm, creating competing objectives~\cite{wei2023jailbroken} routed through a single 1D $J$-space switch~\cite{arditi2024refusal} (the ``red ball'' paradox) rather than deep capability removal.
    \item \textbf{Self-Bootstrapping $\rightarrow$ Collapse \& Escape:} Solving data and R\&D bottlenecks with STaR~\cite{zelikman2022star} and AIDE$^2$~\cite{srikanth2026aide2} enabled models to learn from their own outputs and code, introducing photocopy-style model collapse~\cite{shumailov2024collapse} and evaluation-to-production sandbox escapes~\cite{huggingface2026intrusion}.
\end{itemize}

\subsection{Open Research Gaps (Prospective Directions)}
Synthesizing these 22 works yields five prospective research gaps for future investigations:
\begin{enumerate}
    \item \textbf{Preventing Alignment Regression in Reasoning Agents:} Evaluating multi-turn safety-aware RLHF and representation circuit breakers so LRMs cannot be co-opted as autonomous red-teamers~\cite{hagendorff2026lrm}.
    \item \textbf{Multi-Dimensional Refusal Geometry:} Moving beyond brittle 1D $J$-space directions~\cite{arditi2024refusal} toward distributed, non-linear refusal subspaces resistant to directional ablation and adaptive log-probability search~\cite{andriushchenko2024adaptive}.
    \item \textbf{Verification Gates for Recursive Self-Improvement:} Formalizing cryptographic capability sandboxes and invariant checkers inside RSI loops~\cite{srikanth2026aide2} to prevent evaluation-to-production drift~\cite{huggingface2026intrusion}.
    \item \textbf{Human-to-Synthetic Data Bounds:} Quantifying provenance-verified human data ratios required inside STaR-style and RSI loops~\cite{zelikman2022star,srikanth2026aide2} to prevent distribution tail collapse~\cite{shumailov2024collapse}.
    \item \textbf{Machine-Speed Defensive Telemetry:} Pairing open-weights forensic agents with deterministic cluster containment rules to match 17,600-action autonomous intrusions~\cite{huggingface2026intrusion}.
\end{enumerate}

\section{Conclusion}
Tracing 22 of the most influential papers from 1997 to 2026 reveals a continuous causal chain: solving RNN forgetting with LSTM ``bouncer'' gates gave neural networks memory, Word2Vec turned language into vector math, removing recurrence unlocked Transformer all-to-all attention and data-balanced scaling, scaling produced emergent reasoning alongside brittle alignment, and unaddressed alignment fragility has now converged with autonomous agency and recursive self-improvement. Future progress depends on coupling recursive capability gains with mechanistic, verifiable safety guarantees.

\bibliographystyle{IEEEtran}
\bibliography{references}

\onecolumn
\appendix[Comparative Synthesis Tables]
\label{app:matrix}
This appendix consolidates the cross-era evolutionary summary (Table~\ref{tab:eras}) and the complete 7-column synthesis matrix across all 22 curated sources (Tables~\ref{tab:synth1} and~\ref{tab:synth2}).

\begin{center}
\vspace{6pt}
\captionof{table}{Causal Evolution of Influential LLM Papers (1997--2026)}
\label{tab:eras}
\begin{tabular}{@{}>{\raggedright\arraybackslash}p{2.7cm}>{\raggedright\arraybackslash}p{1.35cm}>{\raggedright\arraybackslash}p{3.4cm}>{\raggedright\arraybackslash}p{3.1cm}>{\raggedright\arraybackslash}p{3.4cm}>{\raggedright\arraybackslash}p{1.65cm}@{}}
\toprule
\textbf{Era (Taxonomy)} & \textbf{Period} & \textbf{Primary Mechanism} & \textbf{Bottleneck Solved} & \textbf{Emergent Limit / Risk} & \textbf{Papers} \\
\midrule
Before 1999 (Pre-Memory) & $<$1999 & CEC \& ``bouncer'' gates & Vanishing gradients (Ch.~1$\rightarrow$9) & Sequential recurrence & \cite{hochreiter1997lstm} \\
Post-1999 Memory & 2013--2016 & Word2Vec math, Seq2Seq, attention & Fixed-vector bottleneck & $O(N)$ sequential wall; adversarial transfer & \cite{mikolov2013word2vec,sutskever2014seq2seq,bahdanau2014nmt,szegedy2013intriguing} \\
Attention Revolution & 2017--2019 & All-to-all attention, MLM & Parallel training; $O(1)$ paths & Quadratic $O(N^2)$ attention cost & \cite{vaswani2017attention,devlin2018bert} \\
Scaling \& RLHF & 2020--2022 & Params+Data laws, CoT, STaR, RLHF & Under-trained params; intent & Superficial safety alignment & \cite{kaplan2020scaling,hoffmann2022chinchilla,brown2020gpt3,wei2022cot,zelikman2022star,ouyang2022instructgpt} \\
Adversarial Fallout & 2023--2024 & GCG, adaptive search, many-shot & Exposing safety failure modes & 100\% ASR; long-context exploits & \cite{wei2023jailbroken,zou2023gcg,andriushchenko2024adaptive,anthropic2024manyshot} \\
Current State (J-Space \& RSI) & 2024--2026 & 1D ``red ball'' refusal, LRM agents, AIDE$^2$ & Automated R\&D; tail collapse & 97.14\% LRM ASR; sandbox escapes & \cite{arditi2024refusal,shumailov2024collapse,hagendorff2026lrm,huggingface2026intrusion,srikanth2026aide2} \\
\bottomrule
\end{tabular}
\end{center}

\begin{center}
\vspace{6pt}
\scriptsize
\setlength{\tabcolsep}{3pt}
\captionof{table}{7-Column Paper Synthesis Matrix: Foundations, Scaling \& RLHF (Papers [1]--[13])}
\label{tab:synth1}
\begin{tabular}{@{}>{\raggedright\arraybackslash}p{2.3cm}>{\raggedright\arraybackslash}p{2.1cm}>{\raggedright\arraybackslash}p{2.5cm}>{\raggedright\arraybackslash}p{2.4cm}>{\raggedright\arraybackslash}p{2.1cm}>{\raggedright\arraybackslash}p{2.3cm}>{\raggedright\arraybackslash}p{2.3cm}@{}}
\toprule
\textbf{Paper} & \textbf{Limitations} & \textbf{Key Findings} & \textbf{Core Contributions} & \textbf{Datasets} & \textbf{Code / GitHub} & \textbf{Future Gaps} \\
\midrule
Hochreiter \& Schmidhuber (1997)~\cite{hochreiter1997lstm} & Synthetic tasks; sequential recurrence & Bridges $>$1,000 steps; $O(1)$ per step & CEC \& multiplicative memory gates & Synthetic long-lag sequences & Not released (1997) & Parallel long-range architectures \\
Mikolov et al. (2013)~\cite{mikolov2013word2vec} & Static vectors; polysemy conflated & 1.6B words in $<$1 day; linear analogies & CBOW \& Skip-gram (Word2Vec) & Google News (1.6B), analogies & Google Code C impl. (later) & Contextualized representations \\
Sutskever et al. (2014)~\cite{sutskever2014seq2seq} & Fixed-vector bottleneck; OOV penalty & 34.8 BLEU (36.5 rerank); reversed input & Deep LSTM Seq2Seq framework & WMT'14 En-Fr & Not publicly released & Variable memory \& OOV handling \\
Bahdanau et al. (2014)~\cite{bahdanau2014nmt} & Retains recurrent encoder/decoder & Soft alignment matches phrase SMT & Additive soft-attention mechanism & WMT'14 En-Fr & Not publicly released & Pure attention without recurrence \\
Szegedy et al. (2013)~\cite{szegedy2013intriguing} & Vision CNNs; no universal defense & Imperceptible perturbations transfer & Adversarial examples & ImageNet, MNIST & Not publicly released & Formal robustness \& transferability \\
Vaswani et al. (2017)~\cite{vaswani2017attention} & Quadratic $O(N^2)$ memory/compute & 28.4/41.8 BLEU in 3.5d on 8 GPUs & Transformer \& Multi-Head Attention & WMT'14 En-De/En-Fr, WSJ & \url{github.com/tensorflow/tensor2tensor} & Sub-quadratic attention \& pre-training \\
Devlin et al. (2018)~\cite{devlin2018bert} & Needs fine-tuning; [MASK] mismatch & SOTA on 11 tasks (GLUE 80.5\%) & MLM \& NSP bidirectional pre-training & BooksCorpus, Wiki, GLUE, SQuAD & \url{github.com/google-research/bert} & Zero/few-shot generative scaling \\
Kaplan et al. (2020)~\cite{kaplan2020scaling} & Overestimated params vs. tokens & Smooth power-law loss over 7+ orders & Empirical compute/param/data laws & WebText2, OpenAI corpora & Not publicly released & Compute-optimal token allocation \\
Hoffmann et al. (2022)~\cite{hoffmann2022chinchilla} & Ignores inference overtraining & 70B (1.4T tokens) beats 280B Gopher & Equal param--token scaling (Chinchilla) & MassiveText, MMLU, BIG-bench & Not publicly released & Inference-optimal \& data-wall limits \\
Brown et al. (2020)~\cite{brown2020gpt3} & NLI struggles; bias \& contamination & 175B learns in-context without grads & Emergent few-shot learning at 175B & CommonCrawl, WebText2, SuperGLUE & Not publicly released & Intent alignment \& safety tuning \\
Wei et al. (2022)~\cite{wei2022cot} & Emerges at $\sim$100B+; manual prompts & 540B beats fine-tuned GPT-3 on GSM8K & CoT intermediate reasoning prompts & GSM8K, SVAMP, StrategyQA & Not publicly released & Automated rationale bootstrapping \\
Zelikman et al. (2022)~\cite{zelikman2022star} & Needs ground-truth verifier & Matches 30$\times$ model on CommonsenseQA & STaR iterative rationale loop & CommonsenseQA, GSM8K & \url{github.com/ezelikman/STaR} & Reward hacking \& collapse control \\
Ouyang et al. (2022)~\cite{ouyang2022instructgpt} & Simple mistakes; labeler bias & 1.3B InstructGPT preferred over 175B & 3-stage SFT $\rightarrow$ RM $\rightarrow$ PPO RLHF & OpenAI API, TruthfulQA & Not publicly released & Adversarial robustness beyond RLHF \\
\bottomrule
\end{tabular}
\end{center}

\begin{center}
\vspace{6pt}
\scriptsize
\setlength{\tabcolsep}{3pt}
\captionof{table}{7-Column Paper Synthesis Matrix: Safety Fallout, $J$-Space \& Agentic Era (Papers [14]--[22])}
\label{tab:synth2}
\begin{tabular}{@{}>{\raggedright\arraybackslash}p{2.3cm}>{\raggedright\arraybackslash}p{2.1cm}>{\raggedright\arraybackslash}p{2.5cm}>{\raggedright\arraybackslash}p{2.4cm}>{\raggedright\arraybackslash}p{2.1cm}>{\raggedright\arraybackslash}p{2.3cm}>{\raggedright\arraybackslash}p{2.3cm}@{}}
\toprule
\textbf{Paper} & \textbf{Limitations} & \textbf{Key Findings} & \textbf{Core Contributions} & \textbf{Datasets} & \textbf{Code / GitHub} & \textbf{Future Gaps} \\
\midrule
Wei et al. (2023)~\cite{wei2023jailbroken} & Early GPT-4/Claude; qualitative & 100\% red-team success; scaling fails & Competing objectives \& mismatched gen. & Red-team evaluation sets & Not publicly released & Safety--capability parity \\
Zou et al. (2023)~\cite{zou2023gcg} & High white-box cost; high perplexity & Vicuna suffixes transfer to GPT/Claude & GCG universal adversarial suffixes & AdvBench & \url{github.com/llm-attacks/llm-attacks} & Low-perplexity \& mechanistic defense \\
Andriushchenko et al. (2024)~\cite{andriushchenko2024adaptive} & Needs logprobs or prefilling API & 100\% ASR on GPT-4o, Claude, R2D2 & Adaptive template + logprob search & HarmBench, JailbreakBench & \url{github.com/tml-epfl/llm-adaptive-attacks} & Hardening APIs \& adaptive defenses \\
Anthropic (2024)~\cite{anthropic2024manyshot} & Requires long context windows & Power-law ASR rise with 100s of shots & Many-Shot Jailbreaking discovery & Synthetic multi-shot dialogues & Not publicly released & Context-invariant alignment \\
Arditi et al. (2024)~\cite{arditi2024refusal} & White-box; open models up to 72B & Ablating 1D direction disables refusal & 1D $J$-space refusal \& ablation & HarmBench, JailbreakBench, Alpaca & \url{github.com/andyrdt/refusal_direction} & Non-linear multi-D refusal \\
Shumailov et al. (2024)~\cite{shumailov2024collapse} & Assumes uncurated replacement & Distribution tails vanish across models & Theory \& proof of Model Collapse & WikiText-2, GMMs, VAEs & Nature Code Availability & Provenance-verified data mixing \\
Hagendorff et al. (2026)~\cite{hagendorff2026lrm} & 70-item benchmark; 4 attacker LRMs & Autonomous LRMs hit 97.14\% ASR & Alignment regression via LRM agents & 70-prompt 7-domain suite & \url{github.com/pe0-p1e/attack-lrm} (repro) & Preventing offensive co-optation \\
Hugging Face (2026)~\cite{huggingface2026intrusion} & Single forensic case study & $\sim$17,600 actions in 4.5d; 0-day \& K8s & Machine-speed intrusion \& GLM-5.2 & Production forensic logs & Not applicable & Sub-second autonomous containment \\
Srikanth et al. (2026)~\cite{srikanth2026aide2} & 8-day run; needs isolated harness & AIDE$^2$ found 7 edits; hack 55\%$\rightarrow$32\% & Closed-loop RSI on agent codebase & FML-Bench \& 4 held-out suites & \url{github.com/AlexWortega/OpenRsi} (RSI) & Safe verification gates for RSI \\
\bottomrule
\end{tabular}
\end{center}

\end{document}